\documentclass{article}

\usepackage[preprint]{neurips_2019}

\usepackage[utf8]{inputenc} 
\usepackage[T1]{fontenc}    
\usepackage{hyperref}       
\usepackage{url}            
\usepackage{booktabs}       
\usepackage{amsfonts}       
\usepackage{nicefrac}       
\usepackage{microtype}      
\usepackage{amsmath}
\usepackage{amssymb}
\usepackage{algorithm,algorithmic}
\usepackage{enumerate}
\usepackage{multirow}
\usepackage{graphicx}
\usepackage{subfigure}
\usepackage{enumitem}
\usepackage{booktabs}
\usepackage{caption}

\title{Efficient EM-Variational Inference for Hawkes Process}

\author{%
  Feng Zhou\\
  Data61, CSIRO\\
  University of New South Wales\\
  \And
  Zhidong Li\\
  University of Technology Sydney\\
  \AND
  Xuhui Fan\\
  University of New South Wales\\
  \And
  Yang Wang\\
  University of Technology Sydney\\
  \AND
  Arcot Sowmya\\
  University of New South Wales\\
  \And
  Fang Chen\\
  University of Technology Sydney\\
  \AND
}

\begin{document}

\maketitle

\begin{abstract}
In classical Hawkes process, the baseline intensity and triggering kernel are assumed to be a \textit{constant} and \textit{parametric function} respectively, which limits the model flexibility. To generalize it, we present a fully Bayesian nonparametric model, namely \textit{Gaussian process modulated Hawkes process} and propose an \textit{EM-variational} inference scheme. In this model, a transformation of Gaussian process is used as a prior on the baseline intensity and triggering kernel. By introducing a latent branching structure, the inference of baseline intensity and triggering kernel is decoupled and the variational inference scheme is embedded into an EM framework naturally. We also provide a series of schemes to \textit{accelerate} the inference. Results of synthetic and real data experiments show that the underlying baseline intensity and triggering kernel can be recovered without parametric restriction and our Bayesian nonparametric estimation is superior to other state of the arts. 
\end{abstract}

\section{Introduction}
\label{introduction}

Point process is a common statistical model in describing the pattern of event occurrence for many real world applications, such as earthquakes~\citep{marsan2008extending} and finance~\citep{hewlett2006clustering}. The influence of past events on future ones is a vital factor in the clustering effect in point process. Many models have been proposed to describe the interaction, such as Hawkes process~\citep{hawkes1971spectra} and correcting model~\citep{ogata1984inference}. Of those models, Hawkes process is the most extensively used one for modelling the \textit{self-exciting} phenomenon. 

Recently, the Hawkes process has been used as an intensity estimator in a wide range of domains such as social networks~\citep{rodriguez2011uncovering}, criminology~\citep{mohler2011self} and financial engineering~\citep{embrechts2011multivariate}. One of the key challenges for Hawkes process is to select the function for baseline intensity and triggering kernel. The vanilla Hawkes process is assumed to be \textit{constant} for the baseline intensity and \textit{parametric function} for the triggering kernel e.g. the exponential decay kernel or power-law kernel. The parametric model introduces convenience to inference but is \textit{inconsistent} with many real applications, e.g. the baseline intensity of civilian deaths due to insurgent activity is changing over time~\citep{lewis2011nonparametric} and the triggering kernel of vehicle collision is a periodic decay function~\citep{zhou2018refined}. To avoid the issue of model selection, various nonparametric models have been proposed based on the gridding domain, for example, modelling the triggering kernel as a histogram function~\citep{marsan2008extending,eichler2017graphical,reynaud2010adaptive}. However, this kind of approximation is limited in that the grid on which to represent the baseline intensity or triggering kernel is arbitrary and we have to tradeoff between precision and computation complexity. To model the baseline intensity and triggering kernel with continuous change, we propose a fully Bayesian nonparametric model for Hawkes process in this paper. We relax any formulated assumptions for both baseline intensity and triggering kernel to model them as smooth functions. The Bayesian priors on both components are a transformation of Gaussian process which guarantees the nonnegativity constraint. In this setting, the inference can be performed without numerical approximation or gridding the domain. To the best of our knowledge, our paper is the first fully Bayesian nonparametric model for Gaussian process modulated Hawkes process. 

However, the model inference is challengeable. In this paper, we apply a \textit{variational inference}~\citep{wainwright2008graphical} scheme to our model. 
There are two \textit{thorny} subjects: First, the baseline intensity is coupled with triggering kernel in the likelihood of Hawkes process, which drastically increases the complexity of performing variational inference. To address this issue, we introduce the \textit{branching structure} to decouple them. The branching structure is a latent variable and estimated via an expectation-maximization (EM) algorithm~\citep{dempster1977maximum}. The variational inference, thus, can be embedded into an EM framework naturally. Second, although \citeauthor{lloyd2015variational} used variational inference for Poisson process, their method is performed by high dimensional numerical optimization which is time consuming let alone embedding it into EM iterations. To address this issue, we provide some \textit{dimensionality reduction} methods and derive a \textit{closed-form} partial derivative to \textit{speed up} the inference. Specifically, we make the following contributions: 

\begin{itemize}[leftmargin=*]
\item The baseline intensity and triggering kernel are both relaxed to be Bayesian nonparametric functions modulated by transformation of Gaussian process. 
\item We introduce a branching structure to make the variational inference feasible. The branching structure is latent so the variational inference needs to be embeded into an EM framework. The complexity of EM-variational (EMV) algorithm is $\mathcal{O}(RN^2)$ over $R$ EM iterations. 
\item We provide accelerating methods and derive the closed-form partial derivative of evidence lower bound (ELBO). As a result, EMV can be drastically accelerated to be practical and efficient. 
\end{itemize}

\section{Related Work}

Due to the flexibility of nonparametric model, the inference of nonparametric Hawkes process has been largely investigated recently, such as estimating the triggering kernel in an EM framework \citep{marsan2008extending}, the estimation method based on the solution of a Wiener-Hopf equation \citep{bacry2016first} relating the triggering kernel with the second order statistics of its counting process and another method consisting of minimizing a quadratic contrast function \citep{eichler2017graphical,reynaud2010adaptive} which assumes the triggering kernel can be decomposed into a discrete vector. 
However, all of them are frequentist algorithms which are based on likelihood only. When the data is sparse, the likelihood based method is prone to be overfitting but the Bayesian method can effectively avoid the problem with a proper prior. 

From Bayesian nonparametric perspective, most related works recently are based on the Gaussian-Cox process. The Gaussian-Cox process is an inhomogeneous Poisson process with a stochastic intensity function modulated by Gaussian process. 
\citet{moller1998log} provided a finite dimensional log-Gaussian prior for the intensity. 
\citet{cunningham2008fast} proposed a model of renewal process with Gaussian process as prior which still requires domain gridding. 
\citet{adams2009tractable} provided a Markov Chain Monte Carlo (MCMC) inference method for the posterior intensity function of a Cox process with a sigmoid Gaussian process prior. \citeauthor{adams2009tractable} augmented the posterior with latent thinning points to make the inference tractable but the complexity is cubic. To improve the efficiency, \citet{samo2015scalable} introduced a small amount of inducing points and proposed a scalable MCMC inference algorithm. The model has a complexity of $\mathcal{O}(N)$ over $N$ data points (given the number of inducing points). \citet{lloyd2015variational} also adopted the idea of inducing points and proposed a variational inference scheme of which the complexity is also $\mathcal{O}(N)$. Recently, \citet{rousseau2018nonparametric} provided a Bayesian nonparametric estimation method for Hawkes process, in which the prior on triggering kernel is based on piecewise constant function or mixture of Beta distributions, whilst the baseline intensity is still constant. 

\section{Hawkes Process}

A Hawkes process is a stochastic process, whose realization is a sequence of timestamps $D=\{t_i \}_{i=1}^N\in[0,T]$. Here, $t_i$ stands for the time of occurrence for the $i$-th event and $T$ is the observation duration of this process. An important way to characterize a Hawkes process is the conditional intensity function that captures the temporal dynamics. The conditional intensity function is defined as the probability of an event occurring in an infinitesimal time interval $[t,t+dt)$ given historical timestamps before time $t$, $\{t_i|t_i<t\}$. 
The specific form of intensity for Hawkes process is
\begin{equation}
\label{2}
\lambda(t)=\mu(t)+\sum_{t_i<t}\phi(t-t_i)
\vspace{-0.2cm}
\end{equation}
where $\mu(t)>0$ is the baseline intensity and $\phi(\tau)$ ($\tau=t-t_i$) is the triggering kernel. In vanilla Hawkes process, $\mu(t)$ is assumed to be constant and $\phi(\tau)$ is a parametric function, e.g. exponential decay function. The summation of triggering kernels explains the nature of self-excitation, i.e. the occurrence of events in the past will increase the intensity of events occurring in the future. Given $\mu(t)$ and $\phi(\tau)$, the likelihood of Hawkes process can be expressed as
\begin{equation}
\label{3}
p(D|\mu(t),\phi(\tau))=\prod_{i=1}^{N}\lambda(t_i)\exp(-\int_0^T\lambda(t)dt).
\vspace{-0.4cm}
\end{equation}

Using Bayesian framework, the posterior $p(\mu(t),\phi(\tau)|D)$ is doubly-intractable,
which is introduced by \citep{adams2009tractable} due to an intractable integral over $t$ in the numerator and another intractable double integral over $\mu(t)$ and $\phi(\tau)$ in the denominator. To solve the doubly-intractable problem without gridding the domain, we propose a variational inference scheme which is embedded into an EM framework. 

\section{Fully Bayesian Nonparametric Hawkes Process}
For Hawkes process, the intensity $\lambda(t)$ is decided by two functions $\mu(t)$ and $\phi(\tau)$. We assume the baseline intensity and triggering kernel are defined as $\mu(t)=f^2(t)$ and $\phi(\tau)=g^2(\tau)$ to achieve the non-negativity where $f(t)$ and $g(\tau)$ are two Gaussian process distributed stochastic functions on the support of $[0,T]$ and $[0,T_\phi]$ ($T_\phi$ is the support of triggering kernel $\phi(\tau)$), respectively. The square transformation function~\citep{flaxman2017poisson, lloyd2015variational} is used and preferred because the inference can be performed in closed form and it keeps the connection between the data and the variational uncertainty. 

To reduce the dimensionality of model inference, we are inspired by the sparse Gaussian process~\citep{titsias2009variational} to introduce some inducing points, which turns out to be useful in our model. $f$ and $g$ are supposed to be dependent on their corresponding inducing points $\mathbf{Z}_f=\{z_f^m\}_{m=1}^{M_f}$ and $\mathbf{Z}_g=\{z_g^m\}_{m=1}^{M_g}$; the function values of $f$ and $g$ at these inducing points are $\mathbf{u}_f$ and $\mathbf{u}_g$ which are stationary Gaussian processes $\mathbf{u}_f\sim \mathcal{N}(\mathbf{0},\mathbf{K}_{z_f z_f})$ and $\mathbf{u}_g\sim \mathcal{N}(\mathbf{0},\mathbf{K}_{z_g z_g})$ (the prior mean is set to 0 without loss of generality). Given a sample $\mathbf{u}_f$ and $\mathbf{u}_g$, $f$ and $g$ are nonstationary Gaussian processes $f|\mathbf{u}_f\sim\mathcal{GP}(v_f(t),\Sigma_f(t,t^\prime))$ and $g|\mathbf{u}_g\sim\mathcal{GP}(v_g(\tau),\Sigma_g(\tau,\tau^\prime))$ with mean and covariance 
\begin{equation}
\label{5}
v_f(t)=\mathbf{k}_{tz_f}\mathbf{K}_{z_f z_f}^{-1}\mathbf{u}_f,\ \ \ 
\Sigma_f(t,t^\prime)=\mathbf{K}_{tt^\prime}-\mathbf{k}_{tz_f}\mathbf{K}_{z_f z_f}^{-1}\mathbf{k}_{z_f t^\prime}
\vspace{-0.4cm}
\end{equation}
\begin{equation}
\label{7}
v_g(\tau)=\mathbf{k}_{\tau z_g}\mathbf{K}_{z_g z_g}^{-1}\mathbf{u}_g,\ \ \ 
\Sigma_g(\tau,\tau^\prime)=\mathbf{K}_{\tau\tau^\prime}-\mathbf{k}_{\tau z_g}\mathbf{K}_{z_g z_g}^{-1}\mathbf{k}_{z_g \tau^\prime}
\end{equation}
where $\mathbf{k}_{tz_f}$, $\mathbf{K}_{t t^\prime}$, $\mathbf{K}_{z_f z_f}$, $\mathbf{k}_{\tau z_g}$, $\mathbf{K}_{\tau \tau^\prime}$ and $\mathbf{K}_{z_g z_g}$ are matrices evaluated using squared exponential kernel with hyperparameters $\theta_0^f$, $\theta_1^f$, $\theta_0^g$ and $\theta_1^g$, $k(x,x')=\theta_0\exp\left(-\frac{\theta_1}{2}\|x-x'\|^2\right)$. So the joint distribution over $D$, $f$, $u_f$, $g$, $u_g$ given $\Theta_f$ and $\Theta_g$ is 
\begin{equation}
\label{10}
\begin{aligned}
p(D,f,&u_f,g,u_g|\Theta_f,\Theta_g)=p(D|\mu(t)=f^2,\phi(\tau)=g^2)
p(f|\mathbf{u}_f,\Theta_f)p(g|\mathbf{u}_g,\Theta_g)p(\mathbf{u}_f|\Theta_f)p(\mathbf{u}_g|\Theta_g),
\end{aligned}
\end{equation}
where $\Theta_f=\{\theta_0^f, \theta_1^f\}$, $\Theta_g=\{\theta_0^g, \theta_1^g\}$. Throughout this paper, $\Theta$ is a general reference to $\Theta_f$ and $\Theta_g$ and the same for $M$, $\mathbf{Z}$, $\mathbf{K}$ and other variables and we often omit conditioning on $\Theta$. 

\section{Inference}
To use variational inference, the ELBO of model evidence needs to be obtained, which means $f$, $u_f$, $g$ and $u_g$ need to be integrated out in \eqref{10}. However, performing this procedure directly is not easy for variational inference because $\mu(t)$ is coupled with $\phi(\tau)$ in the log-likelihood. To ease inference, the branching structure $\mathcal{P}$ of Hawkes process is introduced to decouple $\mu(t)$ and $\phi(\tau)$. Specifically, we introduce a lower-bound $Q(\mu(t),\phi(\tau)|\mu^{(s)}(t),\phi^{(s)}(\tau))$ \citep{lewis2011nonparametric} of the log-likelihood  based on the current parameter estimation $\mu^{(s)}(t),\phi^{(s)}(\tau)$ and
optimize the lower-bound to obtain the
updates for the parameters. See Appendix \ref{app.a} for detailed derivation. The lower-bound
$Q(\mu(t),\phi(\tau)|\mu^{(s)}(t),\phi^{(s)}(\tau))$ is: 
\begin{equation}
\label{12}
\begin{aligned}
&Q(\mu(t),\phi(\tau)|\mu^{(s)}(t),\phi^{(s)}(\tau))\\
&=\underbrace{\vphantom{\sum_{i=2}^N\left[\sum_{j=1}^{i-1}p_{ij}\log\left(\phi(t_i-t_j)\right)\right]}\left[\sum_{i=1}^Np_{ii}\log(\mu(t_i))\right]-\int_0^T\mu(t)dt}_\text{$\mu(t)$ part}+
\underbrace{\sum_{i=2}^N\left[\sum_{j=1}^{i-1}p_{ij}\log\left(\phi(t_i-t_j)\right)\right]-\sum_{i=1}^N\int_{t_i}^{t_i+T_\phi}\phi(t-t_i)dt}_\text{$\phi(\tau)$ part}\\
&\triangleq \log p(D|\mu(t),\mathcal{P}_{ii})+\log p(D|\phi(\tau),\mathcal{P}_{ij}).
\end{aligned}
\end{equation}
where we can see that given the branching structre, the lower-bound of log-likelihood can be decoupled into two independent parts: $\mu(t)$ part and $\phi(\tau)$ part and $p_{ij}$ can be understood as the probability
that the $i$-th event is affected by a previous event $j$ and $p_{ii}$ is the probability
that $i$-th event is triggered by the baseline intensity. Specifically, they can be defined as:
\begin{equation}
\label{27}
    \begin{aligned}
    p_{ij}=\frac{\phi^{(s)}(\tau_{ij})}{\mu^{(s)}(t_i)+\sum_{j=1}^{i-1}\phi^{(s)}(\tau_{ij})}, \ \ \ 
    p_{ii}=\frac{\mu^{(s)}(t_i)}{\mu^{(s)}(t_i)+\sum_{j=1}^{i-1}\phi^{(s)}(\tau_{ij})}.
    \end{aligned}
\end{equation}
\subsection{Baseline Intensity Part}
Consider the $\mu(t)$ part: $\log p(D|\mu(t)=f^2,\mathcal{P}_{ii})$. $\mathcal{P}_{ii}$ means the diagonal entries of $\mathcal{P}$ and $\mathcal{P}_{ij}$ means the others. We integrate out inducing points $\mathbf{u}_f$ using a variational distribution $q(\mathbf{u}_f)=\mathcal{N}(\mathbf{u}_f|\mathbf{m}_f,\mathbf{S}_f)$ over the inducing points where $\mathbf{S}_f$ is positive-semidefinite and symmetric. We use Jensen's inequality to obtain the ELBO of model evidence of $\mu(t)$ part: 
\begin{equation}
\label{13}
\begin{aligned}
&\log p(D|\mathcal{P}_{ii})
=\log\left[\iint p(D|f,\mathcal{P}_{ii})p(f|\mathbf{u}_f)p(\mathbf{u}_f)\frac{q(\mathbf{u}_f)}{q(\mathbf{u}_f)}d\mathbf{u}_fdf\right]\\
&\geq\iint p(f|\mathbf{u}_f)q(\mathbf{u}_f)d\mathbf{u}_f\log p(D|f,\mathcal{P}_{ii})df
+\iint p(f|\mathbf{u}_f)q(\mathbf{u}_f)df\log\left[\frac{p(\mathbf{u}_f)}{q(\mathbf{u}_f)}\right]d\mathbf{u}_f\\
&=\mathbb{E}_{q(f)}\left[\log p(D|f,\mathcal{P}_{ii})\right]-\text{KL}\left(q(\mathbf{u}_f)||p(\mathbf{u}_f)\right)\triangleq\text{ELBO}_{\mu},
\end{aligned}
\end{equation}
where
\begin{equation}
\label{14}
\begin{aligned}
&q(f)=\int p(f|\mathbf{u}_f)q(\mathbf{u}_f)d\mathbf{u}_f=\mathcal{GP}(f|\tilde{v}_f(t),\tilde{\Sigma}_f(t,t^\prime))\\
\tilde{v}_f(t)=\mathbf{k}_{tz_f}\mathbf{K}_{z_f z_f}^{-1}&\mathbf{m}_f,\ \ \ \tilde{\Sigma}_f(t,t^\prime)=\mathbf{K}_{t t^\prime}-\mathbf{k}_{t z_f}\mathbf{K}_{z_f z_f}^{-1}\mathbf{k}_{z_f t^\prime}+\mathbf{k}_{t z_f}\mathbf{K}_{z_f z_f}^{-1}\mathbf{S}_f\mathbf{K}_{z_f z_f}^{-1}\mathbf{k}_{z_f t^\prime}.
\end{aligned}
\end{equation}

$\text{KL}\left(q(\mathbf{u}_f)||p(\mathbf{u}_f)\right)$ is the KL divergence between two Gaussian distributions, so it has an analytical solution. The expectation of log-likelihood over $q(f)$ can be written as
\begin{equation}
\label{16}
\begin{aligned}
&\mathbb{E}_{q(f)}\left[\log p(D|f,\mathcal{P}_{ii})\right]
&=-\int_0^T\left\{\mathbb{E}_{q(f)}^2[f(t)]+\text{Var}_{q(f)}[f(t)]\right\}dt
+\sum_{i=1}^Np_{ii}\mathbb{E}_{q(f)}\left[\log f^2(t_i)\right],
\end{aligned}
\end{equation}
where we utilize $\mathbb{E}(A^2)=\mathbb{E}^2(A)+\text{Var}(A)$. Equation \eqref{16} also has an analytical solution which is introduced in Appendix \ref{app.b}. 

\subsection{Triggering Kernel Part}

Consider the $\phi(\tau)$ part: $\log p(D|\phi(\tau)=g^2,\mathcal{P}_{ij})$. Similarly, we integrate out inducing points $\mathbf{u}_g$ using $q(\mathbf{u}_g)=\mathcal{N}(\mathbf{u}_g|\mathbf{m}_g,\mathbf{S}_g)$ where $\mathbf{S}_g$ is positive-semidefinite and symmetric. The ELBO of model evidence of $\phi(\tau)$ part is 
\begin{equation}
\label{22}
\begin{aligned}
\log p(D|\mathcal{P}_{ij})
&=\log\left[\iint p(D|g,\mathcal{P}_{ij})p(g|\mathbf{u}_g)p(\mathbf{u}_g)\frac{q(\mathbf{u}_g)}{q(\mathbf{u}_g)}d\mathbf{u}_gdg\right]\\
&\geq
\mathbb{E}_{q(g)}\left[\log p(D|g,\mathcal{P}_{ij})\right]-\text{KL}\left(q(\mathbf{u}_g)||p(\mathbf{u}_g)\right)
\triangleq\text{ELBO}_{\phi},
\end{aligned}
\end{equation}

Similarly, $\text{KL}\left(q(\mathbf{u}_g)||p(\mathbf{u}_g)\right)$ has an analytical solution. $q(g)$ is just \eqref{14} with notation $f$ and $t$ replaced by $g$ and $\tau$, respectively. The expectation of log-likelihood over $q(g)$ can be written as
\begin{equation}
\label{24}
\begin{aligned}
\mathbb{E}_{q(g)}\left[\log p(D|g,\mathcal{P}_{ij})\right]&=-\sum_{i=1}^N\int_0^{T_\phi}\left\{\mathbb{E}_{q(g)}^2[g(\tau)]+\text{Var}_{q(g)}[g(\tau)]\right\}d\tau
+\sum_{i=2}^N\sum_{j=1}^{i-1}p_{ij}\mathbb{E}_{q(g)}\left[\log g^2(\tau_{ij})\right],
\end{aligned}
\vspace{-0.1cm}
\end{equation}
where $\tau_{ij}=t_i-t_j$. Equation \eqref{24} can be solved analytically utilizing same method as $\mu(t)$ part. 

\subsection{EM-Variational Framework}

Our algorithm updates the parameters $\mu(t)$ and $\phi(\tau)$ in an EM iterative
manner which guarantees the log-likelihood increasing monotonically.

\textbf{Update for $\mathcal{P}$}: Utilizing equation \eqref{27}.

\textbf{Update for $\mathbf{m}_f^*$, $\mathbf{S}_f^*$ and $\mathbf{m}_g^*$, $\mathbf{S}_g^*$}:
\begin{equation}
\label{28}
\begin{aligned}
\mathbf{m}_f^*,\mathbf{S}_f^*=\operatorname*{argmax}_{\mathbf{m}_f,\mathbf{S}_f}\left(\text{ELBO}_\mu\right),\ \ \ 
\mathbf{m}_g^*,\mathbf{S}_g^*=\operatorname*{argmax}_{\mathbf{m}_g,\mathbf{S}_g}\left(\text{ELBO}_\phi\right).
\end{aligned}
\vspace{-0.4cm}
\end{equation}

\textbf{Update for $\tilde{v}^*_f$, $\tilde{\Sigma}^*_f$ and $\tilde{v}^*_g$, $\tilde{\Sigma}^*_g$}: Utilizing \eqref{14} with $\mathbf{m}_f^*$, $\mathbf{S}_f^*$, $\mathbf{m}_g^*$ and $\mathbf{S}_g^*$. 

\textbf{Update for $\mu(t)$ and $\phi(\tau)$}:
\begin{equation}
\label{29}
\begin{aligned}
\mu(t)=(\tilde{v}^*_f)^2+\tilde{\sigma}^{2*}_f,\ \ \ 
\phi(\tau)=(\tilde{v}^*_g)^2+\tilde{\sigma}^{2*}_g,
\end{aligned}
\vspace{-0.1cm}
\end{equation}
where $\tilde{\sigma}^{2*}_f$ and $\tilde{\sigma}^{2*}_g$ are diagonal entries of $\tilde{\Sigma}^*_f$ and $\tilde{\Sigma}^*_g$. 
\vspace{-0.1cm}

\section{Inference Speed Up}
The na\"ive implementation of the EM algorithm mentioned above is time consuming. The bottleneck is the update for $\mathbf{m}_f^*$, $\mathbf{S}_f^*$ and $\mathbf{m}_g^*$, $\mathbf{S}_g^*$ because we have to perform numerical optimization. Supposing the number of inducing points $\mathbf{u}_f$ is $M_f$, the dimensionality of the search space for optimization over $\mathbf{m}_f$ and $\mathbf{S}_f$ is $M_f+M_f(M_f+1)/2$. This is a large space even when $M_f$ is small (the case of $\mathbf{u}_g$ is the same). We develop \textit{two tricks} to speed up the algorithm: (1) we show that $\mathbf{m}^*$ does not need to be inferred, (2) we derive the closed-form partial derivative of ELBO with respect to (w.r.t.) $\mathbf{S}$ which means we can obtain a local maximum $\mathbf{S}^*$ directly instead of performing numerical optimization. 

The transformation function is $\mu(t)=f^2$ which is not a bijection. For every $\mu(t)$, there will be two positive-negative symmetric $f(t)$'s. The posterior of $f$ can be written as $p(f|D,\mathcal{P}_{ii})\propto p(D|\mu(t)=f^2,\mathcal{P}_{ii})\mathcal{GP}(f|\mathbf{u}_f)\mathcal{N}(\mathbf{u}_f|\mathbf{0},\mathbf{K}_{z_f z_f})$, where it is straightforward to see the likelihood is symmetric, i.e. the likelihood is the same with $f$ and $-f$. For the prior $\mathcal{GP}(f|\mathbf{u}_f)\mathcal{N}(\mathbf{u}_f|\mathbf{0},\mathbf{K}_{z_f z_f})$, we can integrate out $\mathbf{u}_f$ and the marginal distribution over $f$ is still Gaussian with mean $\mathbf{0}$. Therefore, the prior of $f$ is also symmetric. Conclusively, the posterior $p(f|D,\mathcal{P}_{ii})$ is symmetric. By variational inference, we are approximating $p(f|D,\mathcal{P}_{ii})$ by a normal distribution $q(f)=\mathcal{GP}(f|\tilde{v}_f(t),\tilde{\Sigma}_f(t,t^\prime))$ where $\tilde{v}_f(t)=\mathbf{k}_{tz_f}\mathbf{K}_{z_f z_f}^{-1}\mathbf{m}_f$. Apparently, $\mathbf{m}_f^*=\mathbf{0}$ definitely. The same case applies to the $\phi(\tau)$ part as well to obtain $\mathbf{m}_g^*=\mathbf{0}$. This means $\mathbf{m}_f^*$ and $\mathbf{m}_g^*$ can be safely set to $\mathbf{0}$ without inference. 

With the setting of $\mathbf{m}^*=\mathbf{0}$, the update for $\mathbf{m}_f^*$, $\mathbf{S}_f^*$ and $\mathbf{m}_g^*$, $\mathbf{S}_g^*$ becomes the maximization of ELBO over $\mathbf{S}$ only. We derive the closed-form partial derivative of ELBO over $\mathbf{S}$ which is shown in Appendix \ref{app.c}. If $\mathbf{S}_f$ is a symmetric covariance matrix, $\partial \text{ELBO}_\mu/\partial \mathbf{S}_f=\mathbf{0}$ is a nonlinear system consisting of $M_f(M_f+1)/2$ equations which is still slow because of too many equations in the system even when $M_f$ is small. To further accelerate the inference, we assume $q(\mathbf{u}_f)$ is an independent distribution (mean field approximation \citep{bishop2007pattern}) which means $\mathbf{S}_f$ is a diagonal matrix. We derive the simplified partial derivative when $\mathbf{S}_f$ is diagonal in Appendix \ref{app.c}. In the diagonal case, $\partial \text{ELBO}_\mu/\partial \mathbf{S}_f=\mathbf{0}$ is a nonlinear system consisting of $M_f$ equations which can be solved faster. In experiments, we find this assumption does not make much difference when $\mu(t)$ is not a volatile function. The discussion above applies to the $\phi(\tau)$ part as well. Therefore, the final inference algorithm is shown in Algorithm \ref{alg1} (Appendix \ref{app.d1}). 




\subsection{Constant Baseline Intensity}
If $\mu(t)$ is constant $\mu$, there is no need to compute the corresponding nonlinear system and variables of $\mu(t)$ part, e.g. $\mathbf{k}_{t_i z_f}$, $\mathbf{K}_{z_f z_f}$, $\mathbf{\Psi}_f$ and $\partial \text{ELBO}_\mu/\partial \mathbf{S}_f=\mathbf{0}$. $\mu$ could be estimated by $\mu=\sum_{i=1}^Np_{ii}/T$ in each EM iteration. Consequently, it is faster than the general case. The inference algorithm when $\mu$ is constant (Algorithm \ref{alg2}) is shown in Appendix \ref{app.d2}. 

\subsection{Complexity}
Given the number of inducing points, the complexity of $\mu(t)$ part is $\mathcal{O}(N)$ because of the third term in \eqref{16} and that of $\phi(\tau)$ part is $\mathcal{O}(N^2)$ because of the third term in \eqref{24} (we assume all past events have influence on the current one). Therefore, the overall complexity is $\mathcal{O}(RN^2)$ over $R$ iterations. 
In our experiment on a normal desktop (CPU: i7-6700 with 8GB RAM), the na\"ive implementation costs about two hours for $N=205$, 6 inducing points for both $\mathbf{Z}_f$ and $\mathbf{Z}_g$ and 100 EM iterations. Algorithm \ref{alg1} and \ref{alg2} cost about 4 minutes and 2 minutes respectively in the same setting which drastically reduce the running time compared with the na\"ive implementation. 

\section{Hyperparameter Inference}
Given $D$, $\mathbf{Z}_f$, $\mathbf{Z}_g$, $\mathcal{P}$, $\mathbf{S}_f^*$ and $\mathbf{S}_g^*$ in every EM iteration, $\text{ELBO}_\mu$ and $\text{ELBO}_\phi$ are two functions over $\Theta_f$ and $\Theta_g$ respectively. In every iteration, it is straightforward to perform maximization of $\text{ELBO}$ over $\Theta$ using numerical packages. Normally, we update $\Theta$ every 20 iterations. 
Apart from $\Theta$, the only hyperparameter left is the number and location of inducing points. More details about how to choose the number and location of inducing points are introduced in Appendix~\ref{app.f}. 

\section{Experimental Results}
We evaluate the performance of EMV on both synthetic and real-world data
Specifically, we compare our EMV inference algorithm with the following alternatives:
\begin{itemize}[leftmargin=*]
\item \textbf{Gaussian-Cox (GC) process}: a Gaussian process modulated inhomogeneous Poisson process. The inference is performed by the algrithm in \citep{samo2015scalable}. It is only for \textit{real data}.
\item \textbf{RKHS-Cox (RKHSC) process}: an inhomogeneous Poisson process whose intensity is estimated by a reproducing kernel Hilbert space formulation \citep{flaxman2017poisson}. It is only for \textit{real data}. 
\item \textbf{Parametric Hawkes (PH) process}: the vanilla Hawkes process (constant $\mu$ and exponential triggering kernel $\alpha \exp(-\beta(t-t_i))$). The inference is performed by maximum likelihood estimation. 
\item \textbf{Model Independent Stochastic Declustering (MISD)}: the MISD \citep{lewis2011nonparametric} is an EM-based nonparametric algorithm for Hawkes process, where $\mu$ is constant and $\phi(\cdot)$ is a histogram function. We use \textit{MISD-\#} (\textit{\#} is the number of bins) to indicate the corresponding model. 
\item \textbf{Wiener-Hopf (WH)}: it is another nonparametric algorithm for Hawkes process where $\mu$ is constant and $\phi(\cdot)$ is a continuous function. The inference is based on the solution of a Wiener-Hopf equation \citep{bacry2016first}. 
\end{itemize}
We use the following metrics to evaluate the performance
of various methods:
\begin{itemize}[leftmargin=*]
\item \textbf{\textit{LogLik}}: the log-likelihood of test data using the trained model. This is a metric describing the model prediction ability. 
\item \textbf{\textit{EstErr}}: we define the integral of squared error between the learned $\hat{\phi}(\tau)$ ($\hat{\mu}(t)$) and the ground truth as the estimation error. It is only used for \textit{synthetic data}. 
\item \textbf{\textit{Q-Q plot}}: we transform the \textit{real data} timestamps by the fitted model according to the time rescaling theorem \citep{papangelou1972integrability} and generate the quantile-quantile (Q-Q) plot. Generally speaking, Q-Q plot visualizes the goodness-of-fit for different models. 
\item \textbf{\textit{PreAcc}}: Given an event sequence $\{t_1, t_2, ..., t_{i-1}\}$, we may wish to predict the time of next event $t_i$. The $t_i$ has density $P(t_i=t)=\lambda(t)\exp\left(-\int_{t_{i-1}}^t\lambda(s)ds\right)$. The expectation of $t_i$ should be $\mathbb{E}[t_i]=\int_{t_{i-1}}^\infty tp(t_i=t)dt$. The integral in the equations can be estimated by Monte Carlo method. We predict multiple timestamps in a sequence: if the predicted $\Hat{t}_i$ is within $t_i\pm\epsilon$ where $t_i$ is the real timestamp and $\epsilon$ is an error bound, then it is considered to be a correct prediction; or it is a wrong one. The percentage of correct prediction is defined as the prediction accuracy. 
\end{itemize}

\subsection{Experimental Results on Synthetic Data}
In synthetic data experiments, we compare the performance of our EMV inference algorithm with PH, MISD and WH (GC and RKHSC are excluded because they are Poisson process model and cannot provide $\mu$ and $\phi$). Four cases are considered: (1) $\mu=1$ and $\phi(\tau)=1\cdot\exp(-2\tau)$; (2) $\mu(t)=\left\{
    \begin{aligned}
    &1 \ \ (0<t\leq T/2)\\
    &2 \ \ (T/2<t<T)\\
    \end{aligned}
\right.$ and $\phi(\tau)=1\cdot\exp(-2\tau)$; (3) $\mu=1$ and $\phi(\tau)=\left\{
    \begin{aligned}
    &0.25\sin{\tau} \ \ (0<\tau\leq\pi)\\
    &0 \ \ (\pi<\tau<T_\phi)
    \end{aligned}
\right.$; 
(4) $\mu(t)=\sin\left(\frac{2\pi}{T}\cdot t\right)+1 \ \ (0<t<T)$ and $\phi(\tau)=0.3\left(\sin(\frac{2\pi}{3}\cdot\tau)+1\right)\cdot\exp(-0.7\tau) \ \ (0<\tau<T_\phi)$. 

We use the thinning algorithm \citep{ogata1998space} to generate 100 sets of training data and 10 sets of test data with $T=100$ in four cases and use PH, MISD-10, MISD-20, WH and EMV (Algorithm \ref{alg1} for case 2 and 4 and Algorithm \ref{alg2} for case 1 and 3) to perform inference with $T_\phi=6$. The detailed experimental setup (e.g., hyperparameter selection) can be found in Appendix \ref{app.e1}. 

\begin{figure}
\vspace{-0.3cm}
\centering     
\subfigure[]{\label{fig:a}\includegraphics[width=0.24\textwidth]{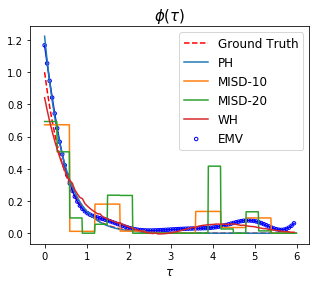}}
\vspace{-0.4cm}
\subfigure[]{\label{fig:b}\includegraphics[width=0.24\textwidth]{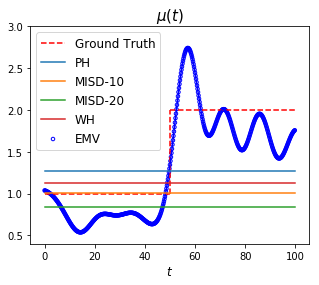}}
\subfigure[]{\label{fig:c}\includegraphics[width=0.24\textwidth]{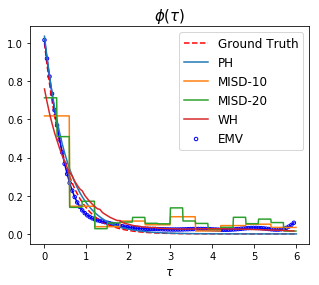}}
\subfigure[]{\label{fig:d}\includegraphics[width=0.24\textwidth]{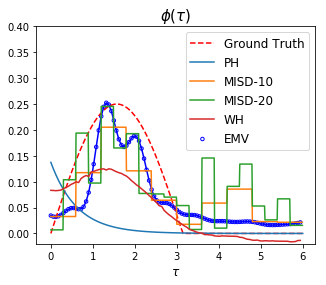}}
\subfigure[]{\label{fig:e}\includegraphics[width=0.24\textwidth]{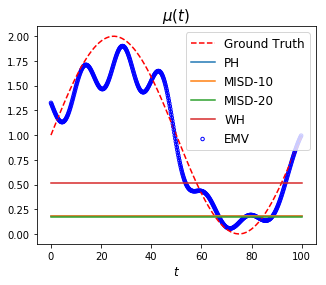}}
\subfigure[]{\label{fig:f}\includegraphics[width=0.24\textwidth]{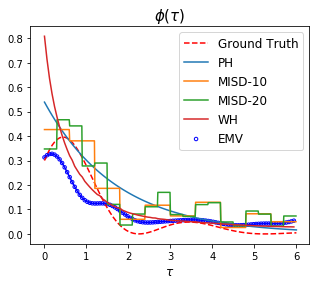}}
\subfigure[]{\label{fig:g}\includegraphics[width=0.25\textwidth]{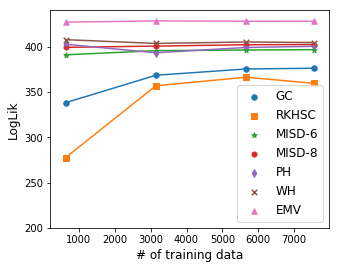}}
\subfigure[]{\label{fig:h}\includegraphics[width=0.25\textwidth]{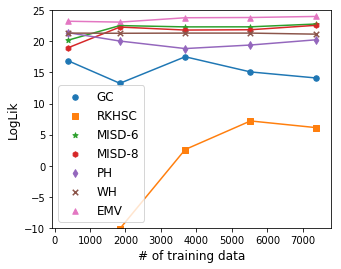}}
\vspace{-0.4cm}
\caption{The synthetic and real data experimental results. (a): the estimated $\hat{\phi}(\tau)$'s in Case 1 (the estimated $\hat{\mu}_{PH}$=0.973, $\hat{\mu}_{MISD-10}$=0.698, $\hat{\mu}_{MISD-20}$=0.620, $\hat{\mu}_{WH}$=0.762, $\hat{\mu}_{EMV}$=0.623); (b) and (c): the estimated $\hat{\mu}(t)$'s and $\hat{\phi}(\tau)$'s in Case 2; (d): the estimated $\hat{\phi}(\tau)$'s in Case 3 (the estimated $\hat{\mu}_{PH}$=1.199, $\hat{\mu}_{MISD-10}$=1.039, $\hat{\mu}_{MISD-20}$=0.861, $\hat{\mu}_{WH}$=1.357, $\hat{\mu}_{EMV}$=1.239); (e) and (f): the estimated $\hat{\mu}(t)$'s and $\hat{\phi}(\tau)$'s in Case 4; (g): the \textit{LogLik} of different models over the number of training data of vehicle collision dataset; (h): the \textit{LogLik} of different models over the number of training data of taxi pickup dataset.}
\label{fig1}
\vspace{-0.7cm}
\end{figure}

The learned $\hat{\mu}(t)$'s and $\hat{\phi}(\tau)$'s are shown in Fig.\ref{fig1}. The \textit{EstErr} and \textit{LogLik} are shown in Tab.\ref{tab1}. \textbf{We can see our EMV algorithm outperforms other alternatives in almost all cases except Case 1. This is because only our EMV algorithm is capable of estimating nonparametric $\mu(t)$ and $\phi(\tau)$ concurrently}; the reason PH is the best in Case 1 is the parametric model assumption just matches with the ground truth which is a rare situation in real applications. 

\begin{table}[htbp]
\vspace{-0.5cm}
\caption{\textit{EstErr} and \textit{LogLik} of synthetic data. $\mu$ is constant in case 1 and 3.}
\label{tab1}
\begin{center}
\scalebox{0.8}{
\begin{tabular}{c|c|c|c|c|c|c}
\hline
\multicolumn{2}{c|}{}   & PH & MISD-10 & MISD-20 & WH & EMV\\ \hline
\multirow{3}{*}{Case 1} & $\textit{EstErr}(\hat{\mu},\mu)$ & \textbf{0.072} & 9.116 & 14.381 & 5.621 & 14.196\\ \cline{2-7} 
                        & $\textit{EstErr}(\hat{\phi}(\tau),\phi(\tau))$ & \textbf{0.008} & 0.075 & 0.106 & 0.009 & 0.015\\ \cline{2-7}
                        & \textit{LogLik} & \textbf{-37.91} & -41.87 & -45.13 & -38.71 & -39.58\\ \hline
\multirow{3}{*}{Case 2} & $\textit{EstErr}(\hat{\mu}(t),\mu(t))$ & 64.362 & 72.847 & 81.008 & 83.883 & \textbf{10.946}\\ \cline{2-7} 
                        & $\textit{EstErr}(\hat{\phi}(\tau),\phi(\tau))$ & 0.015 & 0.043 & 0.058 & 0.013 & \textbf{0.002}\\ \cline{2-7}
                        & \textit{LogLik} & 93.64 & 91.91 & 90.93 & 93.72 & \textbf{96.85}\\ \hline
\multirow{3}{*}{Case 3} & $\textit{EstErr}(\hat{\mu},\mu)$ & 3.960 & \textbf{0.155} & 1.923 & 12.745 & 5.738\\ \cline{2-7}
                        & $\textit{EstErr}(\hat{\phi}(\tau),\phi(\tau))$ & 0.098 & 0.021 & 0.031 & 0.026 & \textbf{0.013}\\ \cline{2-7}
                        & \textit{LogLik} & -70.18 & -51.66 & -51.59 & -53.15 & \textbf{-51.44}\\ \hline
\multirow{3}{*}{Case 4} & $\textit{EstErr}(\hat{\mu}(t),\mu(t))$ & 107.951 & 114.033 & 118.016 & 60.106 & \textbf{10.436}\\ \cline{2-7} 
                        & $\textit{EstErr}(\hat{\phi}(\tau),\phi(\tau))$ & 0.042 & 0.064 & 0.165 & 0.029 & \textbf{0.018}\\ \cline{2-7} 
                        & \textit{LogLik} & 6.34 & 4.13 & 1.18 & 0.74 & \textbf{10.59}\\ \hline
\end{tabular}}
\end{center}
\vspace{-0.7cm}
\end{table}

\subsection{Experimental Results on Real Data}
In the real data section, we apply our EMV algorithm to \textit{two} different datasets and compare the performance with other alternatives. 

\textbf{\textit{Motor Vehicle Collisions in New York City}}
:
The vehicle collision dataset is from New York City Police Department. 
We filter out weekday records in nearly one month (Sep.18th-Oct.13th 2017).
The number of collisions in each day is about 600. Records in Sep.18th-Oct.6th are used as training data and Oct.9th-13th are held out as test data. 

\textbf{\textit{Green Taxi Pickup in New York City}}
:
This dataset includes trip records from all trips completed in green taxis in New York City from January to June 2016. In experiments, the data from Jan.7th to Feb.1st are used as training data and Jan.2nd-6th are held out as test data. In this period, we filter out pickup dates and times for long-distance trips ($>$ 15 miles) since long-distance trips usually have different patterns with short ones. The number of pickups each day is about 400.

More experimental details (e.g., hyperparameter selection) for these two datasets are provided in Appendix \ref{app.e2}. For each method, we evaluate its performance when the number of training data varies. The \textit{LogLik} of EMV and other alternatives are shown in Fig.~\ref{fig:g} and Fig.~\ref{fig:h}. We can observe PH, MISD-6, MISD-8, WH and EMV outperform GC and RKHSC (both are inhomogeneous Poisson processes); this proves the necessity of utilizing Hawkes process to discover the underlying \textit{self-exciting} phenomenon in both datasets. Besides, our EMV algorithm's \textit{consistent} superiority over other Hawkes process inference algorithms (PH, MISD and WH) whose baseline intensity or triggering kernel is too restricted to capture the dynamics proves \textbf{our EMV algorithm can describe $\mu(t)$ and $\phi(\tau)$ in a completely flexible manner which leads to better goodness-of-fit}. 

To further measure the performance, we generate the \textit{Q-Q plot}. 
The perfect model follows a straight line of $y = x$. We compare the inhomogeneous Poisson process, nonparametric Hawkes process with constant $\mu$ and nonparametric Hawkes process with time-changing $\mu(t)$ (our model) in a \textit{Q-Q plot} in Fig.~\ref{tab2}. \textbf{We can observe that our model is generally closer to the straight line, which suggests its better goodness-of-fit than other models}. For prediction task, we measure the \textit{PreAcc} of all alternatives on both datasets. The result is shown in Tab.~\ref{tab2} where \textbf{we can observe that EMV is obviously superior to other alternatives}. More details are introduced in Appendix~\ref{app.e2}.

\begin{table}
\begin{minipage}{0.4\linewidth}
\caption{The \textit{PreAcc} of all alternatives on both real datasets.}\label{tab2}
\centering
\scalebox{0.8}{
\begin{tabular}{c|c|c}
\hline
Dataset & Vehicle Collision & Taxi Pickup \\ \hline
GC      & 17.3\%            & 53.8\%      \\ \hline
RKHSC   & 29.2\%            & 64.0\%      \\ \hline
PH      & 60.6\%            & 67.1\%      \\ \hline
MISD-6  & 67.6\%            & 68.3\%      \\ \hline
MISD-8  & 67.6\%            & 67.9\%      \\ \hline
WH      & 67.3\%            & 67.5\%      \\ \hline
EMV     & \textbf{71.7}\%            & \textbf{70.4}\%      \\ \hline
\end{tabular}}
\end{minipage}
\hfill
\begin{minipage}{0.6\linewidth}
	\centering
	\subfigure{\label{fig2:a}\includegraphics[width=0.4\textwidth]{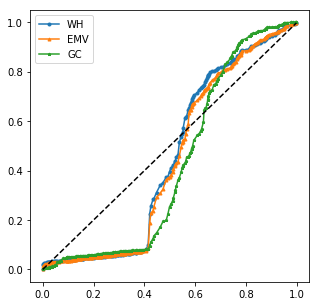}}
    \subfigure{\label{fig2:b}\includegraphics[width=0.4\textwidth]{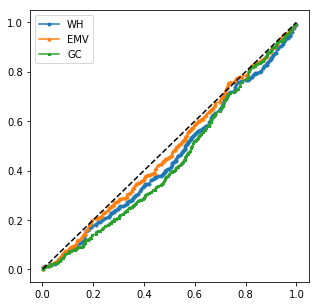}}
	\captionof{figure}{\textit{Q-Q plot} of inhomogeneous Poisson process (GC), nonparametric Hawkes process with constant $\mu$ (WH), nonparametric Hawkes process with time-changing $\mu(t)$ (EMV). Vehicle collision dataset (left), taxi pickup dataset (right).}
	\label{fig2}
\end{minipage}
\vspace{-1cm}
\end{table}

\section{Conclusion}

In vanilla Hawkes process the baseline intensity and triggering kernel are assumed to be a constant and a parametric function respectively, which is convenient for inference but leads to limited capacity for model expression. 
To further generalize the model and perform inference from a Bayesian perspective, we apply the transformation of Gaussian process as prior on the baseline intensity and triggering kernel and solve it with an EM-variational inference algorithm. 
We provide accelerating methods to make the inference efficient. Experiments show that our EM-variational inference can provide better results than the other alternatives. Further investigation directions include the extension to multivariate Hawkes process with sharing properties on the triggering kernels and the more general spatial-temporal process model where the triggering kernel is defined on a multi-dimensional space. 


\bibliography{example_paper}
\bibliographystyle{icml2019}

\newpage
\appendix
\section*{Appendices}
\addcontentsline{toc}{section}{Appendices}
\renewcommand{\thesubsection}{\Alph{subsection}}

\subsection{Proof of Lower-bound}
\label{app.a}
The lower-bound $Q(\mu(t),\phi(\tau)|\mu^{(s)}(t),\phi^{(s)}(\tau))$ in Eq.~\eqref{12} is induced as follows. Based on
Jensen’s inequality, we have
\begin{equation}
\label{15}
\begin{aligned}
&\log p(D|\mu(t), \phi(\tau))=\sum_{i=1}^N \log\left(\mu(t_i)+\sum_{j=1}^{i-1}\phi(t_i-t_j)\right)-\int_0^T\left(\mu(t)+\sum_{t_i<t}\phi(t-t_i)\right)dt\\
&\geq\sum_{i=1}^N\left(p_{ii}\log\frac{\mu(t_i)}{p_{ii}}+\sum_{j=1}^{i-1}p_{ij}\log\frac{\phi(t_i-t_j)}{p_{ij}}\right)-\int_0^T\mu(t)dt-\sum_{i=1}^N\int_{t_i}^{t_i+T_{\phi}}\phi(t-t_i)dt\\
&=\sum_{i=1}^Np_{ii}\log\mu(t_i)-\int_0^T\mu(t)dt+\sum_{i=2}^N\sum_{j=1}^{i-1}p_{ij}\log\phi(t_i-t_j)-\sum_{i=1}^N\int_{t_i}^{t_i+T_\phi}\phi(t-t_i)dt+C\\
&\triangleq \log p(D|\mu(t),\mathcal{P}_{ii})+\log p(D|\phi(\tau),\mathcal{P}_{ij}),
\end{aligned}
\end{equation}
where $C$ is a constant because $p_{ii}$ and $p_{ij}$ are given in the E-step. 

\subsection{Analytical Solution of ELBO}
\label{app.b}
The $\text{KL}\left(q(\mathbf{u}_f)||p(\mathbf{u}_f)\right)$ can be written as
\begin{equation}
\label{11}
\begin{aligned}
\text{KL}\left(q(\mathbf{u}_f)||p(\mathbf{u}_f)\right)=\frac{1}{2}\left[\text{Tr}(\mathbf{K}_{z_f z_f}^{-1}\mathbf{S}_f)+\log\frac{|\mathbf{K}_{z_f z_f}|}{|\mathbf{S}_f|}-M_f+\mathbf{m}_f^T\mathbf{K}_{z_f z_f}^{-1}\mathbf{m}_f\right],
\end{aligned}
\end{equation}
where $\text{Tr}(\cdot)$ means trace, $|\cdot|$ means determinant and $M_f$ is the dimensionality of $\mathbf{u}_f$. 

The first two terms in \eqref{16} have analytical solutions \cite{lloyd2015variational}
\begin{equation}
\label{17}
\int_0^T\mathbb{E}^2_{q(f)}[f(t)]dt=\mathbf{m}_f^T\mathbf{K}^{-1}_{z_f z_f}\mathbf{\Psi}_f\mathbf{K}^{-1}_{z_f z_f}\mathbf{m}_f,
\end{equation}
\begin{equation}
\label{18}
\begin{aligned}
\int_0^T\text{Var}_{q(f)}[f(t)]dt=\theta_0^fT-\text{Tr}(\mathbf{K}^{-1}_{z_f z_f}\mathbf{\Psi}_f)+\text{Tr}(\mathbf{K}_{z_f z_f}^{-1}\mathbf{S}_f\mathbf{K}_{z_f z_f}^{-1}\mathbf{\Psi}_f),
\end{aligned}
\end{equation}
where $\Psi_f(z_f,z_f^\prime)=\int_0^Tk(z_f,t)k(t,z_f^\prime)dt$. For the squared exponential kernel, $\Psi_f$ can be written as \cite{lloyd2015variational} 
\begin{equation}
\label{19}
\begin{aligned}
\Psi_f(z_f,z_f^\prime)=-\frac{(\theta_0^f)^2}{2}\sqrt{\frac{\pi}{\theta_1^f}}\exp\left(-\frac{\theta_1^f(z_f-z_f^\prime)^2}{4}\right)\left[\text{erf}\left(\sqrt{\theta_1^f}(\bar{z}_f-T)\right)-\text{erf}\left(\sqrt{\theta_1^f}\bar{z}_f\right)\right],
\end{aligned}
\end{equation}
where $\text{erf}(\cdot)$ is Gauss error function and $\bar{z}_f=(z_f+z_f^\prime)/2$. 

The third term in \eqref{16} also has an analytical solution \cite{lloyd2015variational}
\begin{equation}
\label{20}
\begin{aligned}
\mathbb{E}_{q(f)}\left[\log f^2(t_i)\right]&=\int_{-\infty}^\infty\log f^2(t_i)\mathcal{N}(f(t_i)|\tilde{v}_f(t_i),\tilde{\sigma}^2_f(t_i))df(t_i)\\
&=-\tilde{G}\left(-\frac{\tilde{v}^2_f(t_i)}{2\tilde{\sigma}^2_f(t_i)}\right)+\log\left(\frac{\tilde{\sigma}^2_f(t_i)}{2}\right)-C,
\end{aligned}
\end{equation}
where $\tilde{\sigma}^2_f(t_i)$ is the diagonal entry of $\tilde{\Sigma}_f(t,t^\prime)$ in \eqref{14} at $t_i$, $C$ is the Euler-Mascheroni constant $0.57721566$ and $\tilde{G}(z)$ is a special case of the partial derivative of the confluent hyper-geometric function $_1F_1(a,b,z)$ \cite{lloyd2015variational}
\begin{equation}
\label{21}
\tilde{G}(z)={_1F_1}^{(1,0,0)}(0,0.5,z).
\end{equation}
However, $\tilde{G}(z)$ does not need to be computed for inference. Actually we only need to know $\tilde{G}(0)=0$ because $\mathbf{m}_f^*=\mathbf{0}$ as we have shown in the section of inference speed up. 

\subsection{Partial Derivative of ELBO}
\label{app.c}
Given $\mathbf{m}_f=\mathbf{0}$, $\text{ELBO}_\mu$ can be written as
\begin{equation}
\label{30}
\begin{aligned}
&\text{ELBO}_\mu=-\left(\theta_0^fT-\text{Tr}(\mathbf{K}^{-1}_{z_f z_f}\mathbf{\Psi}_f)+\text{Tr}(\mathbf{K}^{-1}_{z_f z_f}\mathbf{S}_f\mathbf{K}^{-1}_{z_f z_f}\mathbf{\Psi}_f)\right)\\
&+\sum_{i=1}^Np_{ii}\left(-\tilde{G}(0)+\log(\tilde{\sigma}^2_f(t_i))-\log 2-C\right)\\
&-\frac{1}{2}\left(\text{Tr}(\mathbf{K}_{z_f z_f}^{-1}\mathbf{S}_f)+\log|\mathbf{K}_{z_f z_f}|-\log|\mathbf{S}_f|-M_f\right).
\end{aligned}
\end{equation}

If $\mathbf{S}_f$ is symmetric, $\partial \text{ELBO}_\mu/\partial \mathbf{S}_f$ can be written as
\begin{equation}
\label{31}
\begin{aligned}
&\frac{\partial \text{ELBO}_\mu}{\partial \mathbf{S}_f}=-(2\mathbf{K}_{z_f z_f}^{-1}\Psi_f\mathbf{K}_{z_f z_f}^{-1}-\mathbf{K}_{z_f z_f}^{-1}\Psi_f\mathbf{K}_{z_f z_f}^{-1}\circ\mathbf{I})\\
&+\sum_{i=1}^Np_{ii}\left(2\mathbf{K}^{-1}_{z_f z_f}\mathbf{k}_{z_f t_i}\mathbf{k}_{t_i z_f}\mathbf{K}^{-1}_{z_f z_f}-\mathbf{K}^{-1}_{z_f z_f}\mathbf{k}_{z_f t_i}\mathbf{k}_{t_i z_f}\mathbf{K}^{-1}_{z_f z_f}\circ\mathbf{I}\right)/\tilde{\sigma}_f^2(t_i)\\
&-\frac{1}{2}\left(2\mathbf{K}^{-1}_{z_f z_f}-\mathbf{K}^{-1}_{z_f z_f}\circ\mathbf{I}-(2\mathbf{S}_f^{-1}-\mathbf{S}_f^{-1}\circ\mathbf{I})\right),
\end{aligned}
\end{equation}
where $\mathbf{I}$ means the identity matrix, $\circ$ means Hadamard (elementwise) product and $\tilde{\sigma}_f^2(t_i)=\theta_0^f-\mathbf{k}_{t_i z_f}\mathbf{K}_{z_f z_f}^{-1}\mathbf{k}_{z_f t_i}+\mathbf{k}_{t_i z_f}\mathbf{K}^{-1}_{z_f z_f}\mathbf{S}_f\mathbf{K}^{-1}_{z_f z_f}\mathbf{k}_{z_f t_i}$ is the diagonal entry of $\tilde{\Sigma}_f(t,t^\prime)$ in \eqref{14}. 

If $\mathbf{S}_f$ is diagonal, $\partial \text{ELBO}_\mu/\partial \mathbf{S}_f$ can be further simplified as
\begin{equation}
\label{32}
\begin{aligned}
&\frac{\partial \text{ELBO}_\mu}{\partial \mathbf{S}_f}=-\mathbf{K}_{z_f z_f}^{-1}\Psi_f\mathbf{K}_{z_f z_f}^{-1}\circ\mathbf{I}+\sum_{i=1}^Np_{ii}\frac{\mathbf{K}^{-1}_{z_f z_f}\mathbf{k}_{z_f t_i}\mathbf{k}_{t_i z_f}\mathbf{K}^{-1}_{z_f z_f}\circ\mathbf{I}}{\tilde{\sigma}_f^2(t_i)}-\frac{1}{2}\left(\mathbf{K}^{-1}_{z_f z_f}\circ\mathbf{I}-\mathbf{S}_f^{-1}\right).
\end{aligned}
\end{equation}

Similarly given $\mathbf{m}_g=\mathbf{0}$, $\text{ELBO}_\phi$ can be written as 
\begin{equation}
\label{33}
\begin{aligned}
&\text{ELBO}_\phi=-\sum_{i=1}^N\left(\theta_0^gT_\phi-\text{Tr}(\mathbf{K}^{-1}_{z_g z_g}\mathbf{\Psi}_g)+\text{Tr}(\mathbf{K}^{-1}_{z_g z_g}\mathbf{S}_g\mathbf{K}^{-1}_{z_g z_g}\mathbf{\Psi}_g)\right)\\
&+\sum_{i=2}^N\sum_{j=1}^{i-1}p_{ij}\left(-\tilde{G}(0)+\log(\tilde{\sigma}^2_g(\tau_{ij}))-\log 2-C\right)\\
&-\frac{1}{2}\left(\text{Tr}(\mathbf{K}_{z_g z_g}^{-1}\mathbf{S}_g)+\log|\mathbf{K}_{z_g z_g}|-\log|\mathbf{S}_g|-M_g\right).
\end{aligned}
\end{equation}

If $\mathbf{S}_g$ is symmetric, $\partial \text{ELBO}_\phi/\partial \mathbf{S}_g$ can be written as
\begin{equation}
\label{34}
\begin{aligned}
&\frac{\partial \text{ELBO}_\phi}{\partial \mathbf{S}_g}=-\sum_{i=1}^N(2\mathbf{K}_{z_g z_g}^{-1}\Psi_g\mathbf{K}_{z_g z_g}^{-1}-\mathbf{K}_{z_g z_g}^{-1}\Psi_g\mathbf{K}_{z_g z_g}^{-1}\circ\mathbf{I})\\
&+\sum_{i=2}^N\sum_{j=1}^{i-1}p_{ij}\left(2\mathbf{K}^{-1}_{z_g z_g}\mathbf{k}_{z_g \tau_{ij}}\mathbf{k}_{\tau_{ij} z_g}\mathbf{K}^{-1}_{z_g z_g}-\mathbf{K}^{-1}_{z_g z_g}\mathbf{k}_{z_g \tau_{ij}}\mathbf{k}_{\tau_{ij} z_g}\mathbf{K}^{-1}_{z_g z_g}\circ\mathbf{I}\right)/\tilde{\sigma}_g^2(\tau_{ij})\\
&-\frac{1}{2}\left(2\mathbf{K}^{-1}_{z_g z_g}-\mathbf{K}^{-1}_{z_g z_g}\circ\mathbf{I}-(2\mathbf{S}_g^{-1}-\mathbf{S}_g^{-1}\circ\mathbf{I})\right),
\end{aligned}
\end{equation}
where $\tilde{\sigma}_g^2(\tau_{ij})=\theta_0^g-\mathbf{k}_{\tau_{ij} z_g}\mathbf{K}_{z_g z_g}^{-1}\mathbf{k}_{z_g \tau_{ij}}+\mathbf{k}_{\tau_{ij} z_g}\mathbf{K}^{-1}_{z_g z_g}\mathbf{S}_g\mathbf{K}^{-1}_{z_g z_g}\mathbf{k}_{z_g \tau_{ij}}$ is the diagonal entry of $\tilde{\Sigma}_g(\tau,\tau^\prime)$. 

If $\mathbf{S}_g$ is diagonal, $\partial \text{ELBO}_\phi/\partial \mathbf{S}_g$ can be further simplified as
\begin{equation}
\label{35}
\begin{aligned}
&\frac{\partial \text{ELBO}_\phi}{\partial \mathbf{S}_g}=-\sum_{i=1}^N\mathbf{K}_{z_g z_g}^{-1}\Psi_g\mathbf{K}_{z_g z_g}^{-1}\circ\mathbf{I}+\sum_{i=2}^N\sum_{j=1}^{i-1}p_{ij}\frac{\mathbf{K}^{-1}_{z_g z_g}\mathbf{k}_{z_g \tau_{ij}}\mathbf{k}_{\tau_{ij} z_g}\mathbf{K}^{-1}_{z_g z_g}\circ\mathbf{I}}{\tilde{\sigma}_g^2(\tau_{ij})}\\
&-\frac{1}{2}\left(\mathbf{K}^{-1}_{z_g z_g}\circ\mathbf{I}-\mathbf{S}_g^{-1}\right).
\end{aligned}
\end{equation}

\subsection{EMV Inference Algorithm}
\subsubsection{Time-changing Baseline Intensity}
\label{app.d1}
\begin{algorithm}[H]
\caption{Algorithm for estimation of $\mu(t)$ and $\phi(\tau)$}
\begin{algorithmic}[1]
 \renewcommand{\algorithmicrequire}{\textbf{Input: $\Theta_f$, $\Theta_g$, $\mathbf{Z}_f$, $\mathbf{Z}_g$, $D$, $T$, $T_\phi$}}
 \renewcommand{\algorithmicensure}{\textbf{Output: $\mu(t)$ and $\phi(\tau)$ ($\mathcal{P}$ optional)}}
 \REQUIRE .
 \ENSURE .
 \STATE The precomputation of $\mathbf{k}_{t_i z_f}$, $\mathbf{K}_{z_f z_f}$, $\mathbf{\Psi}_f$, $\mathbf{k}_{\tau_{ij} z_g}$, $\mathbf{K}_{z_g z_g}$ and $\mathbf{\Psi}_g$.
 \STATE Initialize a branching matrix $\mathcal{P}$. 
 \STATE Solve two nonlinear systems $\partial \text{ELBO}_\mu/\partial \mathbf{S}_f=\mathbf{0}$ and $\partial \text{ELBO}_\phi/\partial \mathbf{S}_g=\mathbf{0}$ to get $\mathbf{S}^*_f$ and $\mathbf{S}^*_g$ in diagonal case by \eqref{32} and \eqref{35}. 
 \STATE $q^*(f)$ and $q^*(g)$ can be computed by \eqref{14}.
 \STATE $\mu(t)$ and $\phi(\tau)$ can be computed by \eqref{29}. 
 \STATE Given $\mu(t)$ and $\phi(\tau)$ from the previous step, update $\mathcal{P}$ by \eqref{27}. 
 \STATE Go back to step 3 until the maximum number of iterations. 
 \STATE \textbf{Return} $\mu(t)$ and $\phi(\tau)$ ($\mathcal{P}$ optional). 
\end{algorithmic}
\label{alg1}
\end{algorithm}

\subsubsection{Constant Baseline Intensity}
\label{app.d2}
\begin{algorithm}[H]
\caption{Algorithm for estimation of $\mu$ and $\phi(\tau)$}
\begin{algorithmic}[1]
 \renewcommand{\algorithmicrequire}{\textbf{Input: $\Theta_g$, $\mathbf{Z}_g$, $D$, $T$, $T_\phi$}}
 \renewcommand{\algorithmicensure}{\textbf{Output: $\mu$ and $\phi(\tau)$ ($\mathcal{P}$ optional)}}
 \REQUIRE .
 \ENSURE .
 \STATE The precomputation of $\mathbf{k}_{\tau_{ij} z_g}$, $\mathbf{K}_{z_g z_g}$ and $\mathbf{\Psi}_g$.
 \STATE Initialize a branching matrix $\mathcal{P}$. 
 \STATE Solve the nonlinear system $\partial \text{ELBO}_\phi/\partial \mathbf{S}_g=\mathbf{0}$ to get $\mathbf{S}^*_g$ in diagonal case by \eqref{35}. 
 \STATE After computing $q^*(g)$, $\phi(\tau)$ can be estimated by \eqref{29}. 
 \STATE $\mu=\sum_{i=1}^Np_{ii}/T$
 \STATE Given $\mu$ and $\phi(\tau)$ from the previous step, update $\mathcal{P}$ by \eqref{27}. 
 \STATE Go back to step 3 until the maximum number of iterations. 
 \STATE \textbf{Return} $\mu$ and $\phi(\tau)$ ($\mathcal{P}$ optional). 
\end{algorithmic} 
\label{alg2}
\end{algorithm}

\subsection{Number and Location of Inducing Points}
\label{app.f}
Theoretically, the number $M$ and location $\mathbf{Z}$ of inducing points affect the computation complexity and the estimation quality of $\mu(t)$ and $\phi(\tau)$. If $M$ is too large, the inducing points kernel matrix $\mathbf{K}_{zz}$ will be a high dimensional matrix which leads to high complexity. If $M$ is too small, the inducing points cannot capture the dynamics of $\mu(t)$ or $\phi(\tau)$. 

For a fast inference, we assume the inducing points are uniformly located on the domain. Another advantage of uniform location is that the kernel matrix $\mathbf{K}_{zz}$ has Toeplitz structure which means the matrix inversion can be implemented in $\mathcal{O}(M\log^2M)$ \citep{cunningham2008fast} instead of $\mathcal{O}(M^3)$ in na\"ive implementation. Therefore, the final complexity is reduced from $\mathcal{O}(RN^2M^3)$ to $\mathcal{O}(RN^2M\log^2M)$. 

The number of inducing points depends on the application. If $\mu(t)$ or $\phi(\tau)$ is a volatile function, we need more points to capture the dynamics. In experiments, we can perform preliminary runs: gradually increase the number of inducing points and stop when the resulting $\mu(t)$ or $\phi(\tau)$ is not improved much any more. 

\subsection{Experimental Details}
In this appendix, we elaborate on the details of data generation, processing, hyperparameter setup and some experimental results. 

\subsubsection{Synthetic Data Experimental Details}
\label{app.e1}
The first case is a common one with constant $\mu$ and exponential decay $\phi(\tau)$ and we generate 177 points. For hyperparameters, the bandwidth of WH is set to 0.7 and there are 6 inducing points ($M_g=6$) for EMV. The learned $\hat{\phi}(\tau)$'s are shown in Fig.\ref{fig1}. The \textit{EstErr} and \textit{LogLik} are shown in Tab.\ref{tab1}. The PH is the best in this case because the parametric model assumption matches with the ground truth. 

The second case has a non-constant $\mu(t)$ which is a piecewise constant function. We generate 333 points. The bandwidth of WH is set to 1 and there are 6 inducing points on $\phi(\tau)$ ($M_g=6$) and 8 inducing points on $\mu(t)$ ($M_f=8$) for EMV. The learned results are shown in Fig.\ref{fig1}. The \textit{EstErr} and \textit{LogLik} are shown in Tab.\ref{tab1}. In this case, EMV is the best for both $\mu(t)$ and $\phi(\tau)$ because other alternatives assume $\mu(t)$ is a constant, which is inconsistent with the ground truth. 

The third case has a half sinusoidal triggering kernel. We generate 181 points. The bandwidth of WH is set to 0.5 and there are 10 inducing points ($M_g=10$) for EMV. The learned $\hat{\phi}(\tau)$'s are shown in Fig.\ref{fig1}. The \textit{EstErr} and \textit{LogLik} are shown in Tab.\ref{tab1} in which EMV is still the best for prediction ability although the estimation error is large for $\mu$. The result proves EMV can learn the correct triggering kernel in non-monotonic decreasing cases. 

The fourth case is a general case with time-changing $\mu(t)$ and sinusoidal exponential decay triggering kernel. We generate 212 points. The bandwidth of WH is 0.9 and there are 6 inducing points on $\phi(\tau)$ ($M_g=6$) and 8 inducing points on $\mu(t)$ ($M_f=8$) for EMV. Learned results are shown in Fig.\ref{fig1}. The \textit{EstErr} and \textit{LogLik} are shown in Tab.\ref{tab1} and EMV is still the champion. 

\subsubsection{Real Data Experimental Details}
\label{app.e2}
\textbf{\textit{Vehicle Collision Dataset}}: In daily transportation, car collisions happening in the past will have a triggering influence on the future because of the traffic congestion caused by the initial accident, so there is a self-exciting phenomenon in this kind of application. Hawkes process has already been applied in the transportation domain in the past. However, even using nonparametric Hawkes process like MISD or WH, the baseline intensity is still a constant although the triggering kernel can be relaxed to be nonparametric. This is an inappropriate hypothesis in the vehicle collision application, e.g. the road is quiet at night so the baseline intensity of car accidents is lower than that in the normal time and the traffic is so busy at peak times that the baseline intensity will be increased. Using our EMV inference algorithm, we can learn the time-changing baseline intensity and a flexible triggering kernel simultaneously. 

We compare the performance of EMV, WH, 6-bin MISD (MISD-6), 8-bin MISD (MISD-8), PH, RKHSC and GC. The whole observation period $T$ is set to 1440 minutes (24 hours) and the support of triggering kernel $T_\phi$ is set to 60 minutes. For hyperparameters, the bandwidth of WH is set to 1.2 and there are 6 inducing points on $\phi(\tau)$ ($M_g=6$) and 8 inducing points on $\mu(t)$ ($M_f=8$). The hyperparameters of RKHSC and GC are chosen based on a grid search to minimize the error between the integral of learned intensity and the average number of timestamps on each sequence. The final result is the average of learned $\mu(t)$ or $\phi(\tau)$ of each day. 

We compare the \textit{LogLik} of these algorithms. The result is shown in Fig.~\ref{fig:g}. We can see EMV consistently outperforms other alternatives. This proves the underlying baseline intensity does change over time and EMV algorithm is superior to other alternatives because of the exact description of time-changing $\mu(t)$ and flexible $\phi(\tau)$. 

\textbf{\textit{Taxi Pickup Dataset}}: With the similar setup, we also compare the performance of all methods on the taxi pickup dataset. The whole observation period $T$ is set to 24 hours and the support of triggering kernel $T_\phi$ is set to 1 hour. The time-changing baseline intensity estimated from EMV provides more temporal dynamics information than the other two nonparametric inference algorithms (MISD and WH) where the baseline intensity is constant. The \textit{LogLik} result is shown in Fig.~\ref{fig:h}. The EMV outperforms other alternatives and this proves the superiority of our model. 

\textbf{\textit{Q-Q plot}}:
After the baseline intensity and triggering kernel (or intensity function) have been estimated from the training data series, we can compute the rescaled timestamps of test data:
\begin{equation}
\label{39}
\begin{aligned}
\tau_i=\Lambda(t_i)-\Lambda(t_{i-1}),
\end{aligned}
\end{equation}
where $\Lambda(t_i)=\int_0^{t_i}\lambda(t|\mathcal{H}_{t_i})dt$, $\mathcal{H}_{t_i}$ is the history before $t_i$. According to the time rescaling theorem, $\{\tau_i\}$ are independent exponential random variables with mean $1$ if the model is correct. With further transformation:
\begin{equation}
\label{40}
\begin{aligned}
z_i=1-\exp(-\tau_i),
\end{aligned}
\end{equation}
$\{z_i\}$ are independent uniform random variables on the interval $(0,1)$. Any statistical assessment that measures agreement between the transformed data and a uniform distribution evaluates how well the fitted model agrees with the test data. So we can draw the \textit{Q-Q plot} of the transformed timestamps with respect to the uniform distribution. We randomly select one sequence from the test data of each real dataset. The plots are shown in Fig.~\ref{tab2}. The perfect model follows a straight line of $y = x$. We can observe that our model is generally closer to the straight line, which suggests that EMV
can fit the data better than other models. 


\textbf{\textit{PreAcc}}:For the prediction task, we assume only the top $17\%$ of a sequence is observed ($\epsilon=0.12$ for vehicle collison and $0.24$ for taxi pickup, 500 samples for Monte Carlo integration) and then predict the time of next event, and then the real time of next event is incorporated into the observed data and then predict the further next one and the iteration goes on. Finally, we can calculate the average \textit{PreAcc} of the test data which is shown in Tab.~\ref{tab2}.

\end{document}